\documentclass[10pt,conference]{IEEEtran}

\usepackage[T1]{fontenc}
\usepackage[utf8]{inputenc}

\usepackage{times}
\usepackage{microtype}
\usepackage{booktabs}
\usepackage{multirow}
\usepackage{array}
\usepackage{graphicx}
\usepackage{tabularx}

\usepackage[caption=false,font=footnotesize]{subfig}

\usepackage{amsmath, amssymb, amsfonts}
\usepackage{mathtools}
\usepackage{siunitx}

\usepackage{xcolor}
\usepackage{hyperref}
\usepackage{url}
\usepackage{cite}
\usepackage{enumitem}

\usepackage{algorithm}
\usepackage{algpseudocode}

\usepackage{tikz}
\usetikzlibrary{shapes.geometric, arrows.meta, positioning, fit, calc, backgrounds}

\usepackage{balance}

\definecolor{linkblue}{RGB}{0,82,155}
\hypersetup{
  colorlinks=true,
  linkcolor=linkblue,
  citecolor=linkblue,
  urlcolor=linkblue,
  pdfborder={0 0 0}
}

\definecolor{modalityBg}{RGB}{242, 245, 250}
\definecolor{modalityBorder}{RGB}{180, 190, 210}
\definecolor{coreModelFill}{RGB}{215, 225, 255}
\definecolor{procFill}{RGB}{255, 248, 230}
\definecolor{fuseFill}{RGB}{220, 245, 230}
\definecolor{dataFill}{RGB}{245, 245, 245}

\title{Token-Region Guided Cross-Attention Fusion for Multimodal Affect Interpretation }

\author{
\IEEEauthorblockN{Musa Tur Farazi}
\IEEEauthorblockA{\textit{Department of Computer Science and Engineering} \\
\textit{Bangladesh University of Engineering and Technology}}
\and
\IEEEauthorblockN{Nufayer Jahan Reza}
\IEEEauthorblockA{\textit{Department of Biomedical Engineering} \\
\textit{Bangladesh University of Engineering and Technology}}
}

\begin{document}
\maketitle

\begin{abstract}
Automated analysis of multimodal content on social networks has become a critical task for understanding public sentiment and information diffusion in the digital age. However, classifying internet memes remains computationally challenging due to the intricate interplay between visual cues and embedded, often stylized, text, particularly in low-resource languages like Bengali Language. This paper addresses the detection of political intent in Bengali memes by introducing Multimodal Cross-Attention Fusion framework. We first leverage a Vision-Language Model to extract high-fidelity OCR text from noisy meme images. Subsequently, we encode visual and textual features and synthesize them through a cross-modal multi-head attention mechanism that aligns semantic tokens with visual regions. We also investigate the integration of a domain-specific political lexicon as a knowledge prior. Experimental evaluation on the \textit{PoliMemeDecode}\footnote{Dataset Available at : \href{https://www.kaggle.com/competitions/poli-meme-decode-cuet-cse-fest/data}{Kaggle/PoliMemeDecode}} dataset shows that our attention-based fusion significantly outperforms unimodal baselines and standard concatenation methods, achieving a state-of-the-art Macro-F1 of approximately 0.94. Interpretability analyzes further confirm that the model effectively learns to ground textual semantics in visual evidence.
\end{abstract}

\section{Introduction}
Internet memes have become a popular medium for sharing ideas and humor on social media. Although often lighthearted, they frequently carry serious political content, propaganda, or satire, particularly in the Global South. Detecting political intent in these memes is challenging due to the complex interplay between visual cues and embedded text, which often includes sarcasm, cultural references, and code-mixed languages such as Bengali. Most prior research has focused on high-resource languages, leaving a significant gap in automated understanding tools for low-resource languages.

\begin{figure*}[t]
    \centering
    \includegraphics[width=\linewidth, height=8cm]{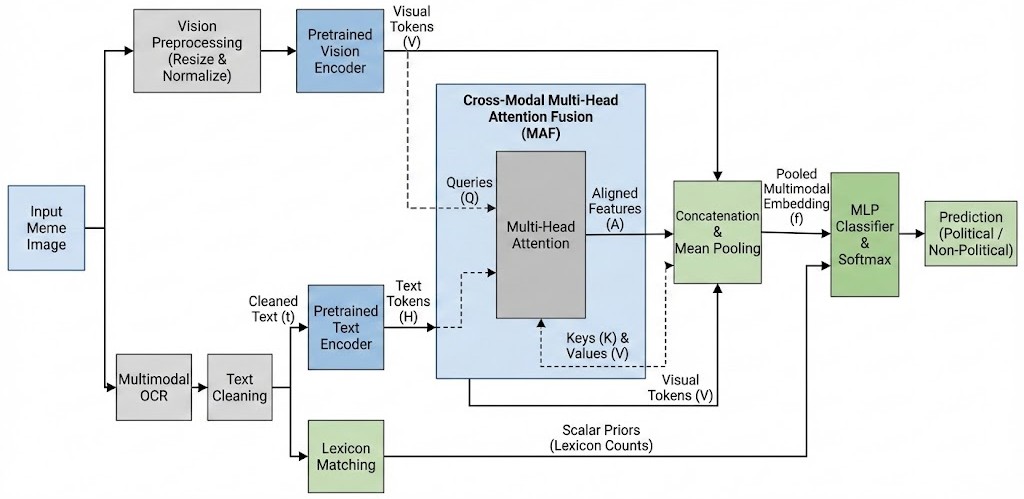}
    \caption{The proposed Multimodal Attention Fusion (MAF) pipeline. The system extracts text using a VLM, encodes image and text via pretrained backbones, aligns them using cross-modal attention, and incorporates lexicon-based domain priors for final classification.}
    \label{fig:pipeline}
\end{figure*}

To address this challenge, we present a practical multimodal pipeline designed to classify Bengali memes as either \textit{Political} or \textit{Non-Political}. As illustrated in Fig.~\ref{fig:pipeline}, our approach operates in three integrated stages. First, we tackle the difficulty of reading stylized meme text using a Vision-Language Model (e.g. Qwen2-VL) for robust OCR extraction, which effectively handles noisy backgrounds and artistic fonts. Second, we encode the raw modalities using powerful pretrained backbones: CLIP extracts high-level visual embeddings, while XLM-RoBERTa captures multilingual textual semantics. Finally, these features are synthesized via our proposed Multimodal Attention Fusion (MAF) module. Unlike simple concatenation, MAF aligns visual and textual tokens through multi-head attention and incorporates a curated political lexicon as a domain prior to guide the classification process.

This work contributes to the field of low-resource affective computing by introducing a novel fusion architecture that specifically addresses the nuances of political meme detection. We demonstrate that moving beyond simple late fusion to a cross-modal attention mechanism significantly improves feature alignment between image and text. Furthermore, we show that integrating lightweight, lexicon-augmented knowledge acts as an effective domain prior, boosting performance in data-scarce regimes. 

\section{Related Work}
Early meme classification research largely targeted English toxic/hateful content. The Hateful Memes Challenge demonstrated that unimodal models are insufficient and established multimodal baselines \cite{kiela2020hateful}. Subsequent work expanded to offensive meme detection, including GNN-based target identification \cite{9582340} and multimodal frameworks such as MOMENTA \cite{pramanick-etal-2021-momenta-multimodal}, often relying on richer external resources in the same line as \cite{sharma2022disarm}.

Recent English-centric advances move beyond simple fusion. \textbf{PromptHate} uses caption-based prompting with RoBERTa and reports strong AUROC/accuracy on HarM \cite{cao2022prompting}. \textbf{ISSUES} applies textual inversion with CLIP and achieves competitive accuracy on the Hateful Memes unseen test set \cite{burbi2023mapping}. For interpretability, \textbf{Thakur et al.} combine CLIP and BERTweet features for example-based reasoning and show multimodal gains on MAMI \cite{thakur2023multimodal}. \textbf{Hamza et al.} fine-tune VisualBERT for religiously hateful memes and improve performance over standard pretrained baselines \cite{hamza2024multimodal}.

In low-resource Bangla, progress is recent. MUTE provided an early Bengali hateful meme dataset \cite{hossain-etal-2022-mute}. Ahsan et al.\ introduced MIMOSA (including a Political category) and proposed attention-based fusion (MAF) \cite{ahsan-etal-2024-multimodal}. Our work follows this direction but targets binary Political vs.\ Non-Political detection and adds a lexicon-based knowledge component.

A key challenge is combining the image and OCR text. Previous works often use late fusion via feature concatenation \cite{kiela2020hateful}\cite{sharma2022disarm}\cite{pramanick-etal-2021-momenta-multimodal}\cite{hossain-etal-2022-mute}, while newer methods model cross-modal interactions with co-/cross-attention, e.g., DORA \cite{hossain2024deciphering} and attentive fusion \cite{ahsan-etal-2024-multimodal}. We adopt multi-head attention to capture token--visual alignment, and we complement learned features with lightweight domain knowledge: a political lexicon signal (cf.\ graph-augmented approaches \cite{9582340}).

Finally, we leverage large pretrained models. CLIP provides strong visual representations \cite{radford2021clip}, but Bengali OCR requires a multilingual text encoder; we pair CLIP with XLM-RoBERTa \cite{conneau-etal-2020-unsupervised}, similar in spirit to low-resource multimodal setups \cite{hossain2024deciphering}, while emphasizing tight attention-based fusion for Bengali political meme detection.

\section{Problem Formulation and Evaluation Protocol}
\label{sec:problem}

\subsection{Task Definition}
Let $\mathcal{D}=\{(x_i,\hat{t}_i,c_i,y_i)\}_{i=1}^{N}$, where $x_i$ is an RGB meme image, $\hat{t}_i$ is cached OCR text, $c_i\in[0,1]$ is OCR confidence (set $c_i{=}1$ if unavailable), and $y_i\in\{0,1\}$ with $1$=Political and $0$=Non-Political. We learn a classifier producing logits $g_\theta(x,\hat{t},c)\in\mathbb{R}^2$ and
\begin{equation}
p_\theta(y \mid x,\hat{t},c)=\text{softmax}\!\big(g_\theta(x,\hat{t},c)\big).
\end{equation}
Prediction is $\hat{y}=\arg\max_{k\in\{0,1\}} p_\theta(y{=}k\mid x,\hat{t},c)$.

\subsection{Evaluation Metrics}
We prioritize Macro-F1 and the Matthews Correlation Coefficient (MCC). For any class $k\in\{0,1\}$, the metrics are defined as:
\par\nobreak\vspace{-5pt}
{\small
\begin{flalign}
\text{Prec}_k &= \frac{TP_k}{TP_k+FP_k}, \qquad
\text{Rec}_k  = \frac{TP_k}{TP_k+FN_k} & \\
F1_k &= \frac{2 \cdot \text{Prec}_k \cdot \text{Rec}_k}{\text{Prec}_k+\text{Rec}_k}, \quad
\text{MacroF1} = \tfrac{1}{2}\sum_{k} F1_k &
\end{flalign}}
\vspace{-10pt}
{\small
\begin{flalign}
\text{MCC} &= \frac{TP \cdot TN - FP \cdot FN}{\sqrt{(TP+FP)(TP+FN)(TN+FP)(TN+FN)}} &
\end{flalign}}
\noindent where $TP, TN, FP, FN$ being calculated with respect to the positive class (Political)
.
\subsection{Cross-Validation}
For fold $k$, $\mathcal{D}^{(k)}_{\text{train}}=\mathcal{D}\setminus\mathcal{D}^{(k)}_{\text{val}}$ with preserved class ratios. We evaluate out-of-fold (OOF) Macro-F1 over $\bigcup_k \mathcal{D}^{(k)}_{\text{val}}$. We ensemble $K$ fold models by averaging probabilities:
\begin{equation}
\bar{p}(y \mid x,\hat{t},c)=\frac{1}{K}\sum_{k=1}^{K} p_{\theta_k}(y \mid x,\hat{t},c).
\end{equation}

\section{Data Processing and Lexicon Prior}
\label{sec:preprocess}

We use raw OCR output $\hat{t}$ and confidence $c$. We obtain clean text via
\begin{equation}
t=\textsc{Clean}(\hat{t}),
\end{equation}
and resize and/or normalize images to the input resolution of the vision backbone (e.g., $224\times224$) with mild augmentation.

\begin{algorithm}[h]
\caption{\textsc{Clean}$(\hat{t})$: OCR Text Normalization}
\label{alg:clean}
\begin{algorithmic}[1]
\Require Raw OCR string $\hat{t}$
\Require Cleaned text $t$
\State $t \leftarrow \text{lowercase}(\hat{t})$
\State Replace line breaks with spaces. Remove URLs and boilerplate. Keep Bengali Unicode with basic punctuation and drop other symbols.
\State Collapse repeated punctuation and whitespace
\State \Return $t$
\end{algorithmic}
\end{algorithm}

We add a lexicon prior using a seed set $\mathcal{L}_0$ and a fold-only expansion. Lexicon matches are:
\begin{equation}
\ell(t)=\sum_{w\in\mathcal{L}}\mathbb{I}[w\subseteq t].
\label{eq:lexcount}
\end{equation}
Expansion uses smoothed log-odds:
\begin{equation}
s(w)=\log\frac{c_1(w)+\alpha}{C_1+\alpha|\mathcal{V}|}-\log\frac{c_0(w)+\alpha}{C_0+\alpha|\mathcal{V}|}.
\label{eq:logodds}
\end{equation}

\begin{algorithm}[h]
\caption{Fold-Safe Lexicon Expansion}
\label{alg:lexexpand}
\begin{algorithmic}[1]
\Require Training texts/labels $\{(t_i,y_i)\}$, seed lexicon $\mathcal{L}_0$, smoothing $\alpha$, top-$K$
\Ensure Expanded lexicon $\mathcal{L}$
\State Compute token counts $c_1(w)$ (Political) and $c_0(w)$ (Non-Political)
\State Compute $s(w)$ via Eq.~\eqref{eq:logodds} for $w\in\mathcal{V}$
\State $\mathcal{L}_{\text{auto}} \leftarrow$ top-$K$ tokens with largest $s(w)$ and $s(w)>0$
\State \Return $\mathcal{L}=\mathcal{L}_0\cup\mathcal{L}_{\text{auto}}$
\end{algorithmic}
\end{algorithm}

\section{Lexicon-Augmented Multimodal Attention Fusion}
\label{sec:model}

Given an image $x$ and cleaned OCR text $t$, we encode vision/text, align them with cross-modal attention, and
append scalar priors $(\ell(t),c)$ for classification.

\subsection{Encoders, Projection, and Fusion}
CLIP yields visual tokens:
\begin{equation}
V=f_{\text{V}}(x)=[v_1,\ldots,v_M], \qquad v_j \in \mathbb{R}^{d_v}.
\end{equation}
XLM-R yields text tokens:
\begin{equation}
H=f_{\text{T}}(t)=[h_1,\ldots,h_L], \qquad h_i \in \mathbb{R}^{d_t}.
\end{equation}
Project to shared dimension $d$:
\begin{equation}
\tilde{v}_j = W_v v_j + b_v, \quad \tilde{h}_i = W_h h_i + b_h,
\end{equation}
yielding $\tilde{V}\in\mathbb{R}^{M\times d}$ and $\tilde{H}\in\mathbb{R}^{L\times d}$.

\subsection{Cross-Modal Multi-Head Attention}
Queries come from text and keys/values from vision:
\begin{equation}
Q = \tilde{H}, \qquad K = \tilde{V}, \qquad V = \tilde{V}.
\end{equation}
Scaled dot-product attention:
\begin{equation}
\text{Att}(Q,K,V) = \text{softmax}\!\left(\frac{QK^\top}{\sqrt{d_k}}\right)V.
\end{equation}
Multi-head attention:
\begin{align}
\text{head}_r &= \text{Att}(QW_r^Q, KW_r^K, VW_r^V), \\
A &= \text{MHA}(Q,K,V) = \text{Concat}(\text{head}_1,\ldots,\text{head}_R)W^O,
\end{align}
where $A \in \mathbb{R}^{L\times d}$. If only a global CLIP embedding is used ($M=1$), replicate it:
\begin{equation}
\tilde{V}_{\text{rep}} = [\tilde{\bar{v}}, \ldots, \tilde{\bar{v}}] \in \mathbb{R}^{L\times d}.
\end{equation}

\subsection{Pooling, Scalars, and Classifier}
Fuse and mean-pool:
\begin{equation}
F_i \;=\; [\tilde{h}_i \,;\, A_i \,;\, \tilde{v}_i] \in \mathbb{R}^{3d},
\qquad i=1,\ldots,L,
\end{equation}
\begin{equation}
\bar{f} \;=\; \frac{1}{L}\sum_{i=1}^{L} F_i \in \mathbb{R}^{3d}.
\label{eq:meanpool}
\end{equation}
Append scalar priors:
\begin{equation}
s \;=\; [\, \phi(\ell(t)) \,;\, c \,] \in \mathbb{R}^{2},
\end{equation}
\begin{equation}
z \;=\; [\,\bar{f}\,;\,s\,] \in \mathbb{R}^{3d+2}.
\end{equation}
Classifier logits:
\begin{equation}
u \;=\; \text{MLP}(\text{LayerNorm}(z)) \in \mathbb{R}^2.
\end{equation}
Optional lexicon boosting for logit prior :
\begin{equation}
\tilde{u}_1 = u_1 + \beta \,\ell(t), \qquad \tilde{u}_0 = u_0.
\label{eq:lexboost}
\end{equation}

\begin{algorithm}[h]
\caption{Forward Pass of Lexicon-Augmented MAF}
\label{alg:forward}
\begin{algorithmic}[1]
\Require Image $x$, raw OCR $\hat{t}$, confidence $c$
\Ensure Class probabilities $p \in [0,1]^2$
\State $t \leftarrow \textsc{Clean}(\hat{t})$ \Comment{Alg.~\ref{alg:clean}}
\State $\ell \leftarrow \sum_{w\in\mathcal{L}} \mathbb{I}[w \subseteq t]$ \Comment{Eq.~\eqref{eq:lexcount}}
\State $V \leftarrow f_{\text{V}}(x),$
$H \leftarrow f_{\text{T}}(t)$
\State Project to shared dimension: $\tilde{V}, \tilde{H}$
\If{$M=1$}
  \State Replicate $\tilde{\bar{v}}$ to length $L$ to form $\tilde{V}_{\text{rep}}$
  \State $A \leftarrow \text{MHA}(Q=\tilde{H},K=\tilde{V}_{\text{rep}},V=\tilde{V}_{\text{rep}})$
\Else
  \State $A \leftarrow \text{MHA}(Q=\tilde{H},K=\tilde{V},V=\tilde{V})$
\EndIf
\State $\bar{f} \leftarrow \frac{1}{L}\sum_{i=1}^{L} [\tilde{h}_i;A_i;\tilde{v}_i]$
\State $z \leftarrow [\bar{f}; \phi(\ell); c]$
\State $u \leftarrow \text{MLP}(\text{LayerNorm}(z))$
\If{lexicon boosting enabled}
  \State $\tilde{u} \leftarrow (u_0,\,u_1 + \beta \ell)$ \Comment{Eq.~\eqref{eq:lexboost}}
\Else \State $\tilde{u} \leftarrow u$
\EndIf
\State $p \leftarrow \text{softmax}(\tilde{u})$
\State \Return $p$
\end{algorithmic}
\end{algorithm}

\subsection{Training Objective and Optimization}
\label{sec:training}

\subsubsection{Weighted Label-Smoothed Cross-Entropy}
\label{sec:loss}
With $C=2$ classes, class weights $w_k$, and smoothing $\varepsilon\in[0,1)$:
\begin{equation}
\tilde{y}_{i,k} = (1-\varepsilon)\,\mathbb{I}[k=y_i] + \frac{\varepsilon}{C}.
\end{equation}
The objective:
\begin{equation}
\mathcal{L}(\theta)
= -\frac{1}{N}\sum_{i=1}^{N} w_{y_i}\sum_{k=1}^{C} \tilde{y}_{i,k}\log p_{i,k}.
\label{eq:loss}
\end{equation}

\subsubsection{Optimization and Regularization}
\label{sec:optim}
We optimize $\mathcal{L}(\theta)$ with AdamW, using cosine LR scheduling with warmup, dropout in the head,
early stopping on validation Macro-F1.

\subsubsection{Cross-Validation Training and Test-Time Ensembling}
\label{sec:cvtrain}
We train one model per stratified fold and compute OOF predictions; test predictions are ensembled by averaging
fold probabilities.

\begin{algorithm}[h]
\caption{Stratified $K$-Fold Training and Inference}
\label{alg:kfold}
\begin{algorithmic}[1]
\Require Dataset $\mathcal{D}$, number of folds $K$
\Ensure OOF predictions $\hat{y}^{\text{OOF}}$, test probabilities $\bar{p}^{\text{test}}$
\State Split $\mathcal{D}$ into stratified folds $\{(\mathcal{D}^{(k)}_{\text{train}},\mathcal{D}^{(k)}_{\text{val}})\}_{k=1}^K$
\For{$k=1$ to $K$}
    \State Build fold-specific lexicon $\mathcal{L}^{(k)}$ using $\mathcal{D}^{(k)}_{\text{train}}$ 
    \State Train model $\theta_k$ on $\mathcal{D}^{(k)}_{\text{train}}$ by minimizing Eq.~\eqref{eq:loss}
    \State Predict on $\mathcal{D}^{(k)}_{\text{val}}$ to fill OOF predictions $\hat{y}^{\text{OOF}}$
    \State Predict probabilities on test set: $p^{(k)}_{\text{test}}(\cdot)$
\EndFor
\State $\bar{p}^{\text{test}} \leftarrow \frac{1}{K}\sum_{k=1}^{K} p^{(k)}_{\text{test}}$
\State \Return $\hat{y}^{\text{OOF}},\bar{p}^{\text{test}}$
\end{algorithmic}
\end{algorithm}

\section{Experiments}
\label{sec:experiments}

\subsection{Experimental Setup}
We evaluate PoliMemeDecode (Political vs.\ Non-Political) using \textbf{stratified 3-fold cross-validation};
all reported metrics are computed from aggregated \textbf{out-of-fold (OOF)} predictions.

Each meme provides an image and OCR text. We optionally include lexicon matches and OCR confidence as scalar features.

Vision uses OpenCLIP ViT-B/16 (LAION2B). Text uses XLM-RoBERTa-large with lower layers frozen (bottom 12 blocks),nd vision fine-tuning is enabled.

We evaluate ablations on image-only (A0), text-only (A1), text+scalars (A2), text+scalars+boost (A3), concat fusion (A4), MAF (A5), and MAF+scalars+boost (A6).

AdamW with warmup (ratio 0.1), LR $1\times 10^{-5}$, weight decay $1\times 10^{-4}$, early stopping (patience 1), and gradient accumulation. We report Macro-F1 and MCC, additionally ROC-AUC, AP, and Brier score.

\subsection{Main Results}
Table~\ref{tab:results} reports mean $\pm$ std across folds. Both modalities are strong (A0 $>$ A1), while multimodal fusion improves performance; \textbf{MAF (A5)} is best. Lexicon boosting (A3 or A6) degrades performance, consistent with
overconfident errors under keyword priors.

\begin{table}[h]
\centering
\caption{Stratified 3-fold CV results (mean $\pm$ std). Best overall is \textbf{A5 (MAF)}.}
\label{tab:results}
\setlength{\tabcolsep}{3pt}
\renewcommand{\arraystretch}{1.05}
\scriptsize
\begin{tabularx}{\columnwidth}{@{}c >{\raggedright\arraybackslash}X c c@{}}
\toprule
\textbf{ID} & \textbf{Model Variant} & \textbf{Macro-F1} & \textbf{MCC} \\
\midrule
A0 & Image Only (OpenCLIP ViT-B/16) & $0.884 \pm 0.016$ & $0.770 \pm 0.030$ \\
A1 & Text Only (XLM-R large) & $0.856 \pm 0.006$ & $0.713 \pm 0.013$ \\
\midrule
A2 & Text + Scalars & $0.861 \pm 0.008$ & $0.724 \pm 0.014$ \\
A3 & Text + Scalars + Boost & $0.831 \pm 0.005$ & $0.663 \pm 0.010$ \\
A4 & Image + Text (Concat) & $0.935 \pm 0.011$ & $0.871 \pm 0.022$ \\
\textbf{A5} & \textbf{Image + Text (MAF)} & $\mathbf{0.940 \pm 0.005}$ & $\mathbf{0.879 \pm 0.011}$ \\
A6 & Full (MAF + Scalars + Boost) & $0.915 \pm 0.010$ & $0.831 \pm 0.020$ \\
\bottomrule
\end{tabularx}
\end{table}

\subsection{Comparison with Prior Multimodal Meme Models}
Table~\ref{tab:sota_comparison} situates our results relative to representative multimodal meme classifiers on their
\emph{original} benchmarks (not directly comparable across datasets). Despite its simplicity, MAF achieves strong
Macro-F1 on Bengali political memes without auxiliary captioning or inversion pipelines.

\begin{table}[h]
\centering
\caption{Performance landscape of representative multimodal meme classifiers (reported on their respective datasets; not directly comparable to PoliMemeDecode).}
\label{tab:sota_comparison}
\setlength{\tabcolsep}{4pt}
\renewcommand{\arraystretch}{1.2}
\scriptsize
\begin{tabularx}{\columnwidth}{@{}l l l c c@{}}
\toprule
\textbf{Method} & \textbf{Reference} & \textbf{Dataset} & \textbf{Metric} & \textbf{Score} \\
\midrule
\multicolumn{5}{@{}l}{\textit{Existing Baselines (English/Multimodal)}} \\
VisualBERT & Hamza et al. (2024) & Religious Hate & Acc & 70.60 \\
Example-based & Thakur et al. (2023) & MAMI (Misogyny) & F1 & 0.701 \\
ISSUES (CLIP+Inv) & Burbi et al. (2023) & Hateful Memes & Acc & 77.70 \\
PromptHate & Cao et al. (2022) & HarMeme & Acc & 84.47 \\
\midrule
\multicolumn{5}{@{}l}{\textit{Proposed Method (Bengali)}} \\
\textbf{MAF (Ours)} & This Work & \textbf{PoliMemeDecode} & \textbf{Macro-F1} & \textbf{0.940} \\
\bottomrule
\end{tabularx}
\end{table}

\section{Ablation Study and Analysis}
\label{sec:ablation}

We complement quantitative ablations with diagnostics to explain \textit{why} attention-based fusion helps and
\textit{when} heuristics fail.

\subsection{Impact of Modality and Fusion}
\begin{figure}[h]
    \centering
    \includegraphics[width=\columnwidth,height=0.38\textheight,keepaspectratio]{\detokenize{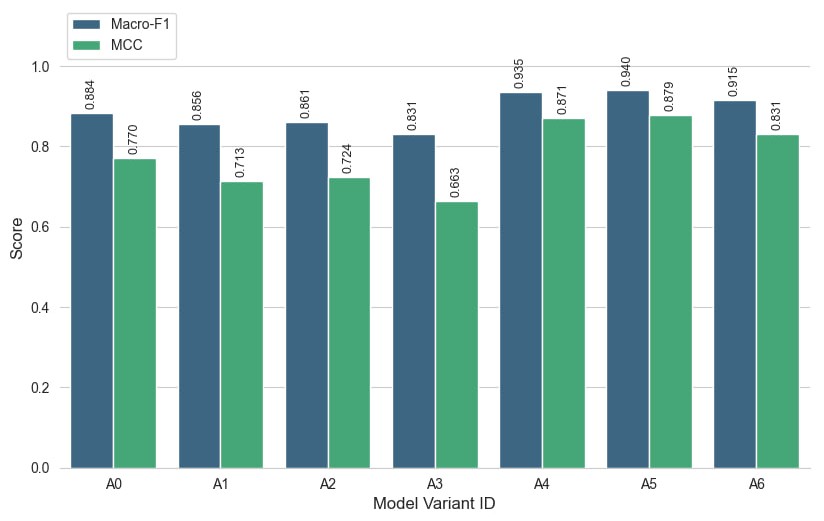}}
    \caption{Ablation performance (Macro-F1 and MCC). Multimodal fusion (A4/A5) yields large gains over unimodal baselines. MAF (A5) provides the strongest overall trade-off.}
    \label{fig:ablation_metrics}
\end{figure}

Figure~\ref{fig:ablation_metrics} summarizes Macro-F1 and MCC across variants; gains largely reflect simultaneous
reductions in FP and FN rather than error trade-offs.

\subsection{Confusion Matrix Analysis (Best Model: A5)}
\begin{figure}[h]
    \centering
    \includegraphics[width=0.8\columnwidth,height=0.2\textheight,keepaspectratio]{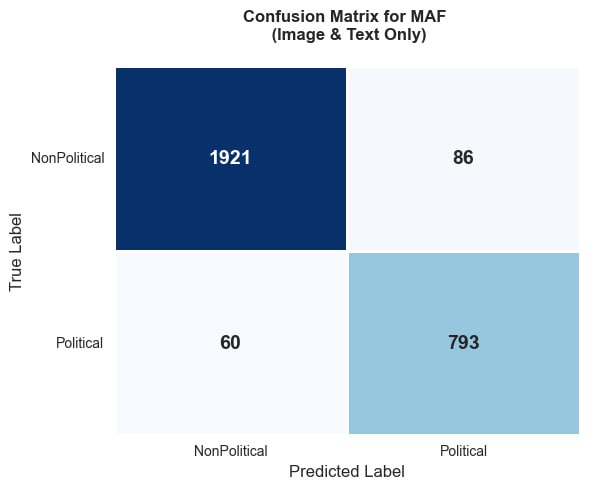}
    \caption{Confusion matrix for A5 (MAF).}
    \label{fig:cm_maf}
    \vspace{-6pt}
\end{figure}

Figure~\ref{fig:cm_maf} shows the confusion matrix of A5. With TN=1921, FP=86, FN=60, TP=793, political recall is
$793/(793+60)=0.930$ and precision is $793/(793+86)=0.902$, indicating a strong balance of errors.

\subsection{Representation Structure (t-SNE)}
\begin{figure}[h]
    \centering
    \includegraphics[width=0.90\columnwidth,height=5cm,keepaspectratio]{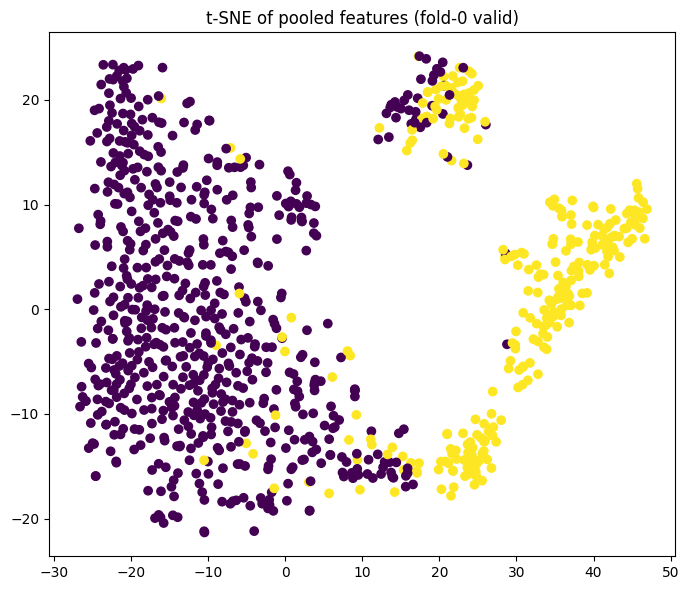}
    \caption{t-SNE of pooled features (fold-0 validation). Political and non-political samples form largely separable regions with a small overlap band.}
    \label{fig:tsne}
    \vspace{-6pt}
\end{figure}

Figure~\ref{fig:tsne} visualizes pooled features (fold-0 validation). Classes form largely separable regions with a
narrow overlap; misclassifications cluster near this boundary, consistent with genuinely ambiguous memes.

\subsection{Calibration and Reliability}
\begin{figure}[h]
    \centering
    \includegraphics[width=0.96\columnwidth,height=0.23\textheight,keepaspectratio]{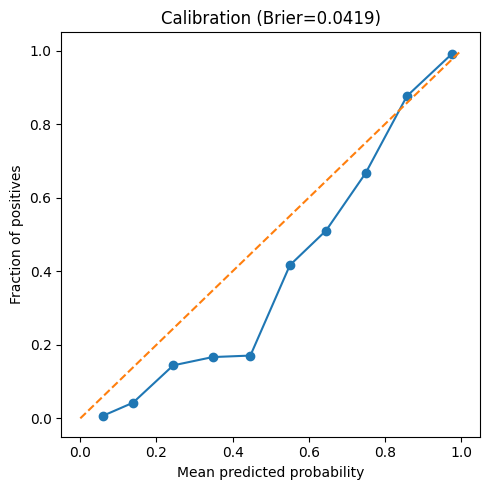}
    \caption{Calibration curve for A5 (Brier=0.0419).}
    \label{fig:calibration}
    \vspace{-6pt}
\end{figure}

Figure~\ref{fig:calibration} evaluates calibration for A5 (Brier=0.0419). The  curve stays close to the
diagonal, with only mild deviation.

\subsection{Role of Lexicon Matches and OCR Confidence}
\begin{figure}[h]
    \centering
    \subfloat[Predicted probability vs.\ lexicon matches\label{fig:lexicon_scatter}]{%
        \includegraphics[width=\columnwidth, height=4.6cm]{\detokenize{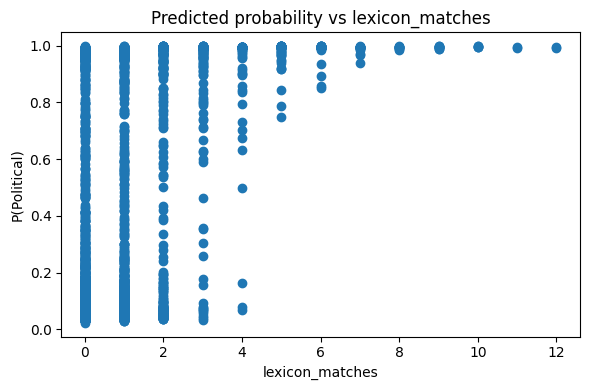}}%
    }%
    \\
    \subfloat[Predicted probability vs.\ OCR confidence\label{fig:ocr_scatter}]{%
    \includegraphics[width=\columnwidth,height=4.6cm]{\detokenize{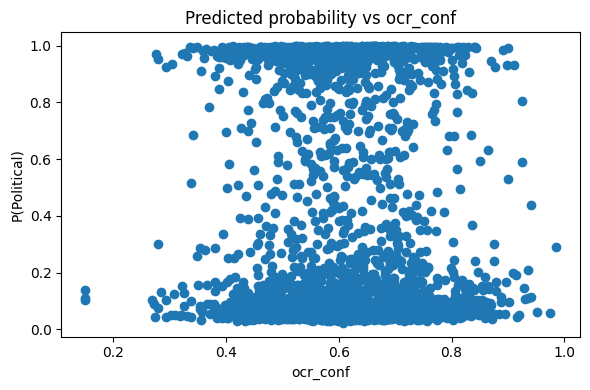}}%
    }
    \caption{Relationship between auxiliary scalars and model confidence.}
    \label{fig:scalar_analysis}
\end{figure}

Figure~\ref{fig:scalar_analysis} relates auxiliary scalars to model confidence. Lexicon matches correlates
with higher political probability, but the wide spread at low counts shows predictions are not purely keyword-based.

\subsection{Visual Interpretability with SmoothGrad}
\begin{figure}[h]
    \centering
    \includegraphics[width=\columnwidth,height=0.40\textheight,keepaspectratio]{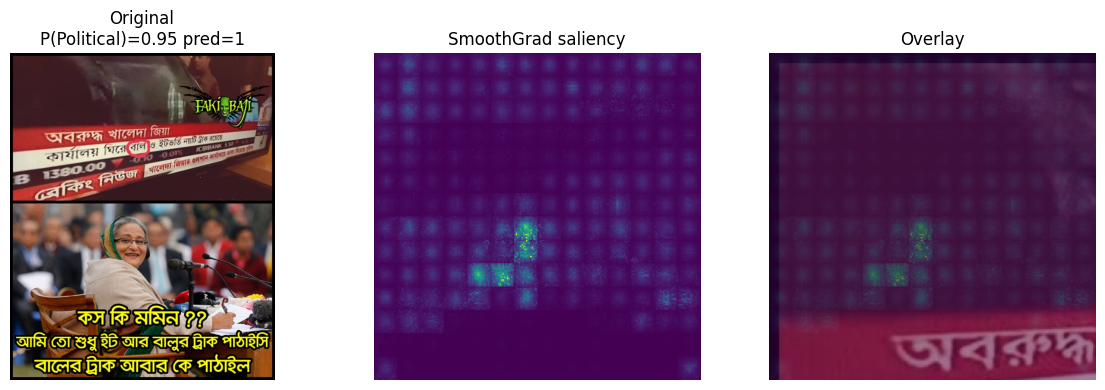}
    \caption{SmoothGrad saliency (A5). Saliency concentrates on overlaid text and salient identity regions}
    \label{fig:saliency}
\end{figure}

Figure~\ref{fig:saliency} shows SmoothGrad saliency: the model focuses on overlaid text and salient identity regions
(e.g., faces/emblems) rather than backgrounds, supporting semantically grounded visual use.

\subsection{OCR Token Contribution Analysis}
\begin{figure}[h]
    \centering
    \includegraphics[width=\columnwidth,height=5cm]{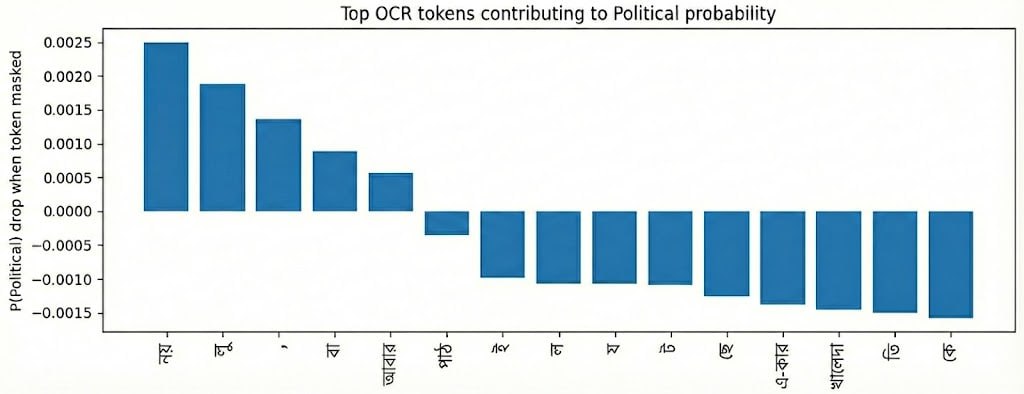}
    \caption{Token-occlusion attribution on OCR text. Bars show the average change in predicted Political probability when a token is masked/removed. Positive values indicate tokens that increase Political probability; negative values indicate tokens that suppress it.}
    \label{fig:token_occlusion}
    \vspace{-6pt}
\end{figure}

We quantify OCR token influence by masking token $w$ and measuring the expected probability drop:
\begin{equation}
\Delta(w) \;=\; \mathbb{E}\big[p(y{=}1 \mid x,t)\;-\;p(y{=}1 \mid x, t_{\setminus w})\big],
\end{equation}
where $t_{\setminus w}$ removes (or masks) $w$. Figure~\ref{fig:token_occlusion} reports the highest-magnitude
positive/negative tokens. XLM-R uses subword units, so attributions are at the fragment level.

\subsection{Why Lexicon Boosting Degraded Performance}
Manual boosting adds a fixed prior term:
\begin{equation}
\label{eq:lexboost_analysis}
\mathbf{z}' = \mathbf{z} + \beta \,\ell(t),
\end{equation}
which can increase recall when explicit political keywords appear, but also increases false positives in neutral or
metaphorical contexts and worsens calibration relative to A5.
\section{Discussion}

\paragraph{Why multimodal attention helps.}
MAF consistently improves over concatenation (A4) and strengthens Macro-F1/MCC, suggesting better alignment between
salient OCR tokens and visual evidence, particularly for context-dependent or sarcastic memes.

\paragraph{Feature engineering vs.\ learned fusion.}
Manual lexicon boosting (A3/A6) reduces performance and calibration, indicating brittle behavior when political
keywords appear in neutral/metaphorical contexts. Learned fusion better conditions lexical cues on the image.

\paragraph{Role of OCR quality.}
Accurate OCR remains important for stylized Bengali and code-mixed text; improved transcripts reduce ambiguity and
enable stronger multimodal reasoning.


\section{Conclusion}
We studied Bengali political meme detection with a multimodal framework that jointly leverages meme imagery and OCR text. Our proposed \textbf{Multimodal Attention Fusion (MAF)} aligns token-level textual evidence with visual cues via cross-modal multi-head attention, and we further explored lightweight domain signals through a political lexicon prior. On the PoliMemeDecode dataset, MAF consistently outperformed unimodal baselines and simple concatenation, achieving the best overall performance with \textbf{Macro-F1 $\approx 0.94$} and \textbf{MCC $\approx 0.88$} under stratified cross-validation. Extensive analyses (confusion matrix, calibration, t-SNE, saliency, and token-occlusion) indicate that the model learns complementary cross-modal evidence rather than relying on brittle keyword triggers, and they explain why manual lexicon boosting can harm calibration and increase false positives. Overall, our results highlight that attention-based fusion of strong pretrained backbones (OpenCLIP + XLM-R) is an effective and practical approach for low-resource, script-diverse political meme understanding.

\balance

\bibliographystyle{IEEEtran}
\bibliography{references}

\end{document}